\documentclass[letterpaper, 10 pt, conference]{ieeeconf}  

\IEEEoverridecommandlockouts                              

\usepackage{amsmath}
\usepackage{amsfonts}
\usepackage{graphicx}
\usepackage{booktabs}
\usepackage{mathtools}
\usepackage{bm}
\usepackage{tabularx}
\usepackage{caption,subcaption}
\usepackage{cite}
\usepackage{xcolor}
\usepackage{float} 
\usepackage{array,multirow}
\usepackage{todonotes}
\usepackage{xspace}  

\newcommand{\dogkernel}{DoG-based kernel\xspace}  

\title{\LARGE \bf
Bayesian Optimization Using Domain Knowledge on the ATRIAS Biped  }

\author{Akshara Rai*$^{1}$, Rika Antonova*$^2$, Seungmoon Song$^1$, William Martin$^1$, Hartmut Geyer$^1$, Christopher G. Atkeson$^1$ 
\thanks{*Both authors contributed equally.}
\thanks{$^1$Robotics Institute, School of Computer Science, Carnegie Mellon University, USA, \texttt{\{arai, seungmoo, wmartin\}@andrew.cmu.edu}, \texttt{\{hgeyer, cga\}@cs.cmu.edu}}
\thanks{$^2$Robotics, Perception and Learning, CSC, KTH Royal Institute of Technology, Stockholm, Sweden, \texttt{antonova@kth.se}}
\thanks{This research was supported in part by National Science Foundation grant IIS-1563807, the Max-Planck-Society, \& the Knut and Alice Wallenberg Foundation. Any opinions, findings, and conclusions or recommendations expressed in this material are those of the author(s) and do not necessarily reflect the views of the funding organizations.}}

\begin{document}

\maketitle
\thispagestyle{empty}
\pagestyle{empty}

\begin{abstract}
Controllers in robotics often consist of expert-designed heuristics, which can be hard to tune in higher dimensions. It is typical to use simulation to learn these parameters, but controllers learned in simulation often don't transfer to hardware. This necessitates optimization directly on hardware. However, collecting data on hardware can be expensive. This has led to a recent interest in adapting data-efficient learning techniques to robotics. One popular method is Bayesian Optimization (BO), a sample-efficient black-box optimization scheme, but its performance typically degrades in higher dimensions. We aim to overcome this problem by incorporating domain knowledge to reduce dimensionality in a meaningful way, with a focus on bipedal locomotion. In previous work, we proposed a transformation based on knowledge of human walking that projected a 16-dimensional controller to a 1-dimensional space. In simulation, this showed enhanced sample efficiency when optimizing human-inspired neuromuscular walking controllers on a humanoid model. In this paper, we present a generalized feature transform applicable to non-humanoid robot morphologies and evaluate it on the ATRIAS bipedal robot -- in simulation and on hardware. We present three different walking controllers; two are evaluated on the real robot.
Our results show that this feature transform captures important aspects of walking and accelerates learning on hardware and simulation, as compared to traditional BO.
\end{abstract}


\section{Introduction}
Locomotion controllers often involve expert-designed heuristics, for example feedback control of the Center of Mass (CoM) and designing reference trajectories. State of the art work in walking robots featuring heuristics includes \cite{feng2015optimization}, \cite{kuindersma2016optimization} and \cite{hubicki2016walking}. These heuristics consist of sets of inter-dependent parameters, which can be hard to tune, especially in higher dimensions. 
This complexity motivates methods for learning parameters automatically. A simple approach is to learn in simulation and deploy on hardware. However, due to differences between simulation and hardware, such as modelling errors, parameters often do not transfer well. On the other hand, directly learning on hardware can require a prohibitive number of samples, making it nearly impossible to learn these controllers using traditional methods. This has led to a surge in interest in data-efficient learning techniques for robotics. 

One popular data-efficient method for learning controller parameters is Bayesian Optimization (BO). BO is a sample-efficient gradient-free black-box optimization method that has been applied to a wide range of robotics problems. For example, \cite{Calandra2016}, \cite{marco2017virtual}, \cite{cully2015robots} try to learn parameters directly on hardware using BO. However, the performance of BO degrades in high dimensions (see~\cite{localBO17} for related discussion). We aim to overcome this problem by incorporating domain knowledge into BO.
\begin{figure}
    \centering
    \includegraphics[width = 0.4\textwidth]{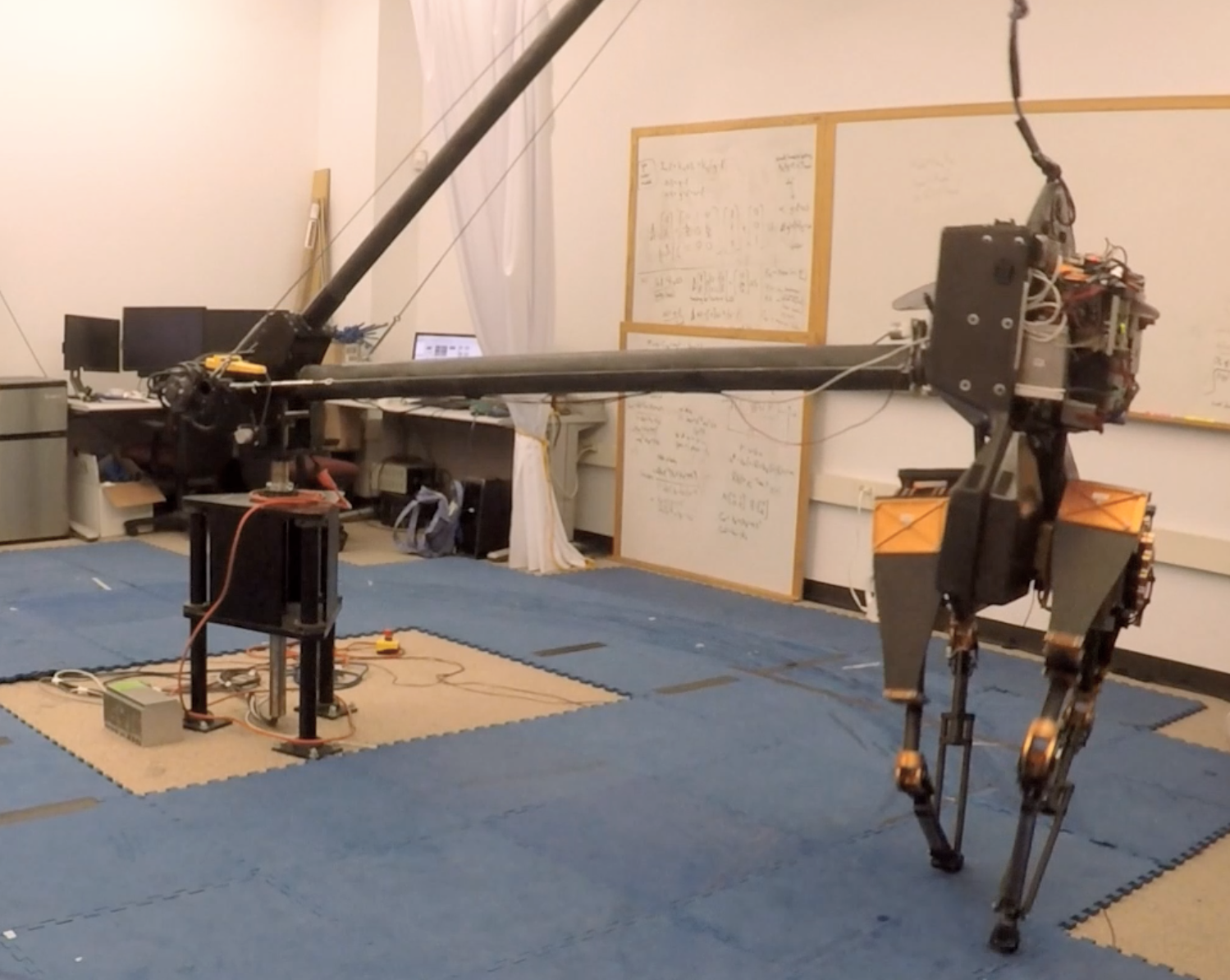}
    \caption{\small{Our testbed is CMU's ATRIAS robot. 
    }}
        \vspace{-5mm}
    \label{fig:atrias}
\end{figure}
In our previous work, we proposed a transformation based on domain knowledge that reparameterized human-like walking controllers based on their behaviour in a high-fidelity simulation.
The idea of using simulation performance to speed up optimization on hardware has been explored in literature before. A common approach is to learn controllers in simulation, and use this as a starting point on hardware. A domain expert would then typically have to fine-tune parameters on hardware. \cite{mordatch2015ensemble}~learn parameters using an ensemble of simulations to account for model uncertainty. \cite{ha2015reducing}~iteratively learn both the model and controller parameters using the differences between simulated behaviors and hardware experiments.
 \cite{marco2017virtual}~use evaluations from simulation as a noisy prior for the optimization on hardware. \cite{cully2015robots}~pre-select high performing controllers from simulation and search among them on hardware. 
 
 In this paper, we generalize our previous human-inspired feature transform to include other system morphologies and controllers. We also present evaluations of our method on the ATRIAS biped robot (Figure \ref{fig:atrias}), for the first time. We evaluate our feature transform on three different controllers -- two of them on hardware. We successfully optimize parameters for a 5-dimensional and 9-dimensional controller on the ATRIAS hardware in less than 10 trials, which proves to be challenging for traditional BO. 
 Our results show that this feature transform extracts useful information from simulations, and leads to an effective transfer of knowledge to hardware. 
We also optimize parameters for a 50-dimensional controller in simulation and obtain promising results. This motivates future work for using our approach on hardware for the 50-dimensional controller as well.
The rest of the paper is organized as follows: In Section \ref{sec:background} we present background on the concepts used in this paper and summarize related work. In Section \ref{sec:dog} we describe our approach of using a locomotion feature transform in detail. Section \ref{sec:atrias_cont} describes our test platform ATRIAS and the controllers used in our experiments. In Section \ref{sec:experiments} we describe our simulation and hardware experiments. Section \ref{sec:conclusions} concludes with further discussion.

\section{Background}
\label{sec:background}

\subsection{Bayesian Optimization and Gaussian Processes}

\begin{figure}[t]
\centering
\begin{subfigure}[t]{0.44\textwidth}
\includegraphics[width=1.0\textwidth]{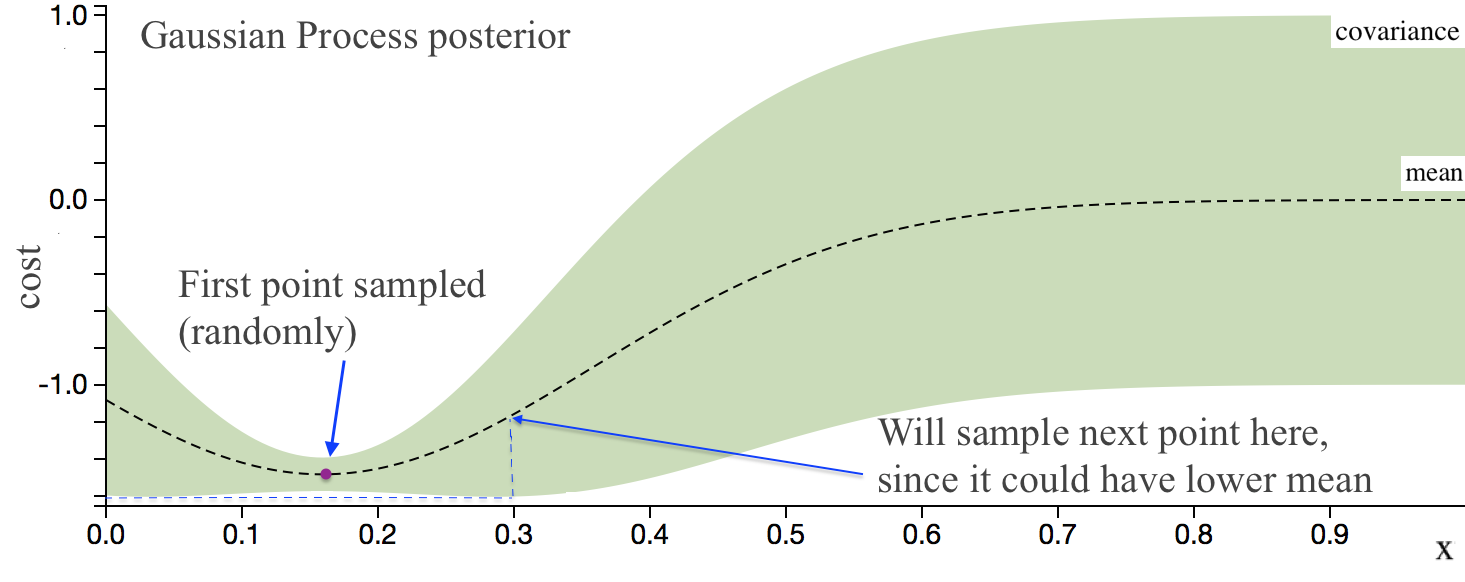}
\end{subfigure}
\vspace{4px}
\begin{subfigure}[t]{0.44\textwidth}
\includegraphics[width=1.0\textwidth]{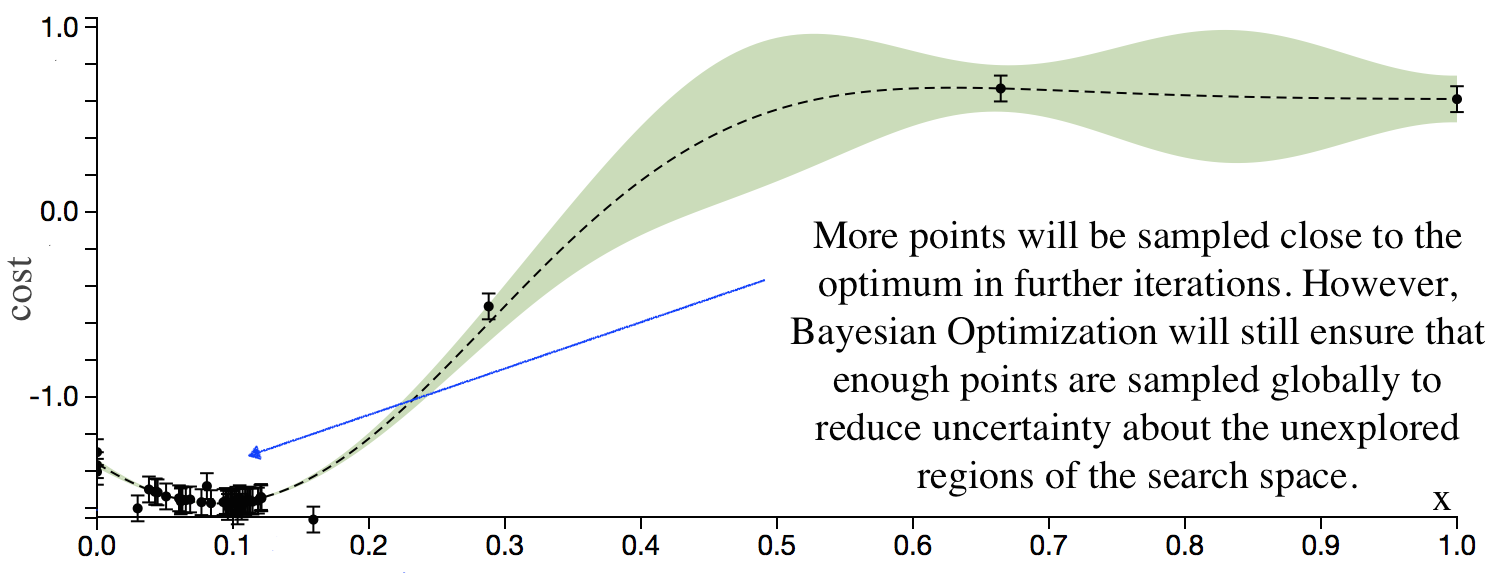}
\end{subfigure}
\caption{\small{Bayesian Optimization posterior for a 1D function. $\ \ \ $ Acquisition function computes the location of points to sample, taking into account both estimated mean and variance (uncertainty).}}
        \vspace{-5mm}
\label{fig_bo_illustration}
\end{figure}

Bayesian Optimization (BO) is a framework for sample-efficient black-box and gradient free global search. 
Recent tutorials ~\cite{BOtutorial2016} and~\cite{BOtutorial2010} provide a  comprehensive overview. 
The goal of BO is to find $\pmb{x}^*$ that optimizes an objective function $f(\pmb{x})$, while executing as few evaluations of $f$ as possible. 
The optimization starts with a prior (which could be uninformed) roughly capturing the prior uncertainty over the value of $f(\pmb{x})$ for each $\pmb{x}$ in the domain. Then an auxiliary optimization function, called \textit{acquisition function}, is used to sequentially select points $\pmb{x}_n$, and $f(\pmb{x}_n)$ is evaluated. The aim of the acquisition function is to automatically balance exploration and exploitation: select points for which the posterior estimate of the objective $f$ is promising, while also decreasing the uncertainty about $f$. An example of BO is shown in Figure \ref{fig_bo_illustration}. 

The prior/posterior mean and variance of $f$ is often expressed by a Gaussian Process (GP):
\begin{align*}
\vspace{-5px}
f(\pmb{x}) \sim \mathcal{GP}(\mu(\pmb{x}), k(\pmb{x}_i, \pmb{x}_j)),
\end{align*}
with mean function $\mu$ and kernel $k$. The prior mean function $\mu(\pmb{x})$ can be set to $0$ if no relevant domain-specific information is available. The kernel $k(\pmb{x}_i, \pmb{x}_j)$ encodes how similar $f$ is expected to be for two inputs $\pmb{x}_i, \pmb{x}_j$. 
The value of $f(\pmb{x}_i$) has a significant influence on the posterior value of $f(\pmb{x}_j)$ if $k(\pmb{x}_i, \pmb{x}_j)$ is large. The Squared Exponential (SE) kernel is a widely used similarity metric:
\begin{equation*}
\vspace{-2px}
k_{SE}(\pmb{x}_i, \pmb{x}_j) = \sigma_k^2 \exp\Big(- \frac{1}{2 \pmb{l}^2} \|\pmb{x}_i - \pmb{x}_j\|^2 \Big),
\end{equation*}
where $\sigma_k^2, \ \pmb{l}^2$ denote signal variance and a vector of length scales respectively; $\sigma_k^2, \ \pmb{l}^2$ are called `hyperparameters' in BO literature. It is customary to adjust these automatically during optimization to learn the overall signal variance and how quickly $f$ varies in each input dimension. 

\subsection{Utilizing trajectory data with Behaviour-based Kernels}

Bayesian Optimization has been used to optimize controllers in various robotics domains. Recent examples from locomotion, mobile robotics and manipulation include~\cite{marco2017virtual, cully2015robots, calandra17thesis, andersson2016model, englert2016combined}. The advantage of using a global search framework like BO is that globally optimal policies can be discovered. These may perform better than human hand-designed policies as well as locally optimal solutions. As shown in~\cite{englert2016combined}, the alternative of using Reinforcement Learning (RL) algorithms for policy search might not discover global optima reliably.

However, the performance of BO degrades in higher dimensions, and the need to improve sample-efficiency arises. One principled approach is to use custom kernels. For example, \cite{wilson2014using} proposed a Behavior Based Kernel (BBK) that uses similarity of trajectories induced by the evaluated policies. A kernel constructed to use behavior-based similarity offered an improvement over a standard SE kernel. However, BBK required obtaining trajectory data every time kernel values $k(\pmb{x}_i, \pmb{x}_j)$ were evaluated. This would not be tractable when optimizing locomotion controllers. The authors suggest combining BBK with a model-based approach to overcome this difficulty. But the question of how to build a useful model while relying only on data from very few trials remains open.

Our approach constructs an informed kernel that incorporates behavior-based similarity, in a manner that ensures $k(\pmb{x}_i, \pmb{x}_j)$ can be obtained efficiently when running BO. We pre-calculate locomotion-specific features on a high-fidelity simulator by running short simulations, enabling us to efficiently calculate the kernel distances during optimization. 

\subsection{Bayesian Optimization for Locomotion Controllers}

There has been work on BO for mobile robots, but these experiments typically involve simpler robots. \cite{lizotte2007automatic} use AIBO quadrupeds, \cite{tesch} use snake robots, \cite{cully2015robots} use hexapods and \cite{calandra2016bayesian} use a small biped.  While quadrupeds, hexapods and snake robots can do dynamic gaits, the percentage of time spent dynamically is small as compared to the time spent statically stable. On the other hand, ATRIAS is a highly dynamic robot due to its point feet, and it cannot be statically stable, except in double-stance on a boom. This makes the optimization harder, as the system is unstable leading to discontinuities in cost functions.

\cite{calandra2016bayesian} use BO for optimizing gaits of a 4 dimensional controller on a small biped. They report needing 30-40 samples for finding walking gaits for a finite-state-machine based controller. Optimizing a higher-dimensional controller, needed for more complex robots, might present even more challenges. The learning could be especially difficult if a significant number of the points/parameters sampled would lead to unstable gaits and falls. Such samples might result in eventual wear and breakage of the robot hardware (even if care is taken to prevent actual falls). Hence, there is a need to either limit search spaces to ``safe" points, or bias the search towards such points. 

\cite{cully2015robots} tabulate best performing points in simulation versus their average score on a behavioural metric for a hexapod robot. This metric then guides BO to quickly find behaviours that can compensate for damage of the robot. The search on hardware is conducted in behaviour space, and limited to pre-selected ``successful” points from simulation.  This helps make their search faster and safer.   However, if an optimal point was not pre-selected,  BO cannot sample it during optimization, losing global optimality guarantees.  ``Best points" are cost-specific, so the map needs to be re-generated for each cost.

Our proposed method generalizes to highly dynamic behaviours and discontinuous cost functions, while maintaining the global guarantees of BO. We also bias our search towards sampling points successful in simulation (but not exclusively), leading to a sample-efficient and safer search.

\section{A general locomotion distance metric}
\label{sec:dog}
In this section, we describe our proposed bipedal locomotion specific feature transform. This transform is designed to generalize to a range of locomotion controllers and robot morphologies, unlike our previous work \cite{rai2016sample}, which focused on human-like robots and controllers.
\subsection{The Determinants of Gait Transformation}
The proposed locomotion feature transform is a generalization of the Determinants of Gaits (DoG) used by physiotherapists to evaluate the quality of human walking \cite{inman1953major}. It consists of the following four walking metrics calculated per step:

\begin{enumerate}
    \item {\bf{$M_1$ : Swing leg retraction}} -- We look at the swing leg trajectory in each swing and if the maximum leg retraction is more than a threshold, we set $M_1 = 1$. Otherwise, $M_1 = 0$.
    \item {\bf{$M_2$ : Center of Mass height}} -- We look at the Center of Mass (CoM) height at the start and end of each step. If the CoM height stays about the same (change is below a threshold), we set $M_2 = 1$. Otherwise, $M_2 = 0$. This metric checks that the robot is not falling across steps, but allows changes in CoM height within a step. 
    \item {\bf{$M_3$ : Trunk lean}} -- We compare the mean trunk lean at the start and end of a step, and if the average lean is about the same (change is below a threshold), we set $M_3 = 1$. Otherwise, $M_3 = 0$. This ensures that the trunk is not changing orientation between steps, but allows the lean to change within a step.
    \item {\bf{$M_4$ : Average walking speed}} -- We evaluate the average speed of a controller per step and set $M_4 = v_{avg}$. Unlike the other metrics, $M_4$ is not binary and helps distinguish between controllers that satisfy all conditions of $M_{1-3}$.
\end{enumerate}

The step metrics $M_{1-4}$ are collected per step $i$ and summed over the total number of steps $N$. 
\begin{align}
\label{eq:dog}
 score^i = \sum_{j=1}^4 M^i_j  \\
 score_{total} = \sum_i^{N} score^i
\end{align}

\begin{figure}
    \centering
    \includegraphics[width=0.45\textwidth]{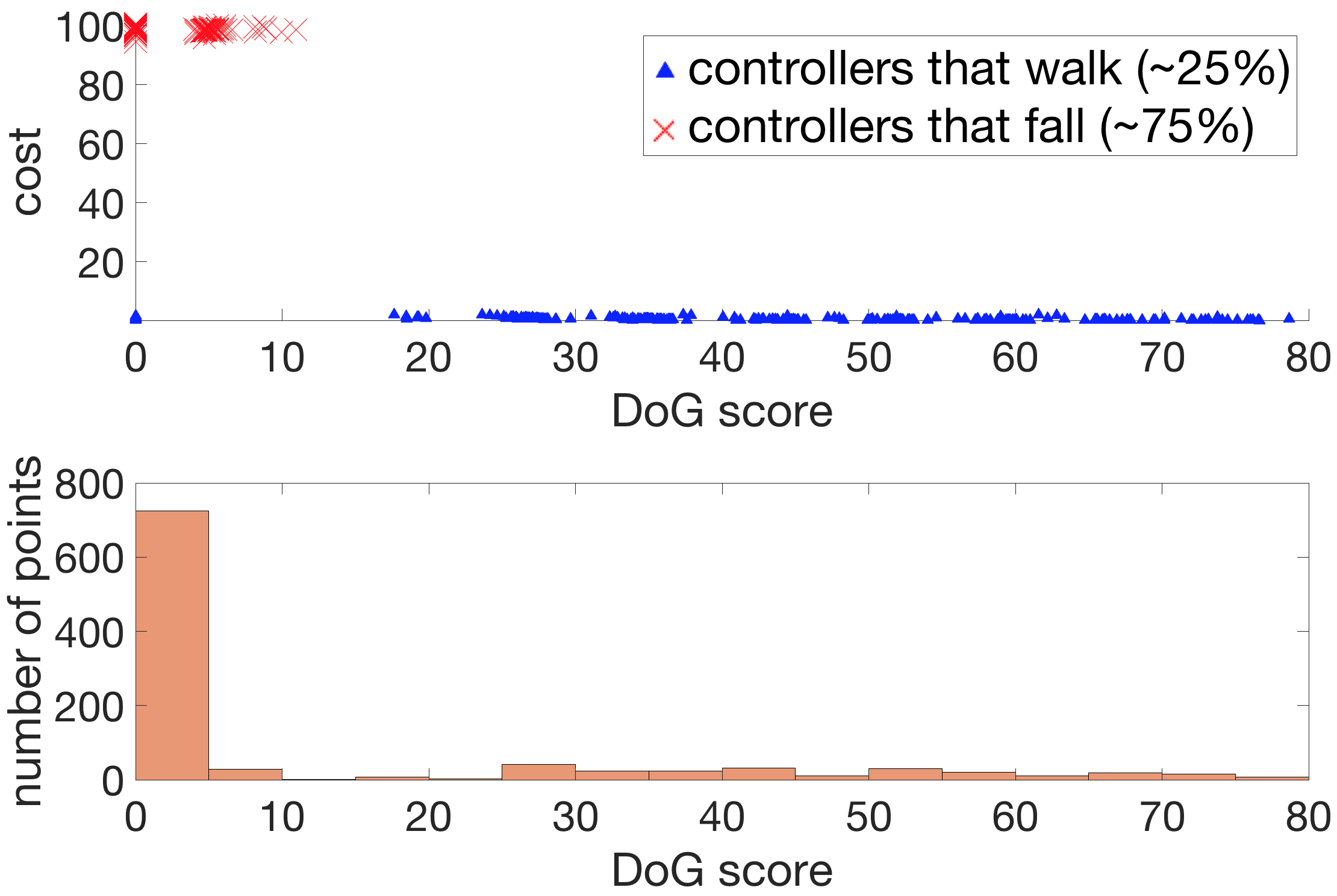}
    \caption{\small{DoG score vs cost for 1000 randomly selected controller parameters (controller \& cost from Sections~\ref{sec:raibert_cont},\ref{sec:exps_sim}). Lower DoG scores usually lead to higher costs \& falling. A few points that step very fast (chatter)  don't fall in simulation, so can have low cost and low DoG score. But such points are very likely to fall on hardware.}}
            \vspace{-5mm}
    \label{fig:dogvsfit}
\end{figure}

In general, a higher score implies better performance of a controller for $M_{1-3}$. If $M_{1-3}$ are 0 for a particular controller, it is likely to fall. On the other hand if $M_{1-3}$ are 1 for a controller, it is likely to walk. However, the score doesn't have a fixed relationship to a particular cost function. The cost depends on the specific desired behavior/outcome.

Controllers that chatter (step very fast, with step time less than $100ms$) can have a large number of steps before falling. Since this could lead to a misleadingly high score, the DoG score is scaled by the fraction of time the simulation walked before falling. If the simulation terminated at time $t_{sim}$ and the desired time for simulation was $t_{max}$, the final DoG score $\phi$ becomes:
\begin{equation}
    \phi = score_{total} \cdot \frac{t_{sim}}{t_{max}}
\end{equation}
The DoG score helps cluster controllers based on their behaviour in simulation. The hope is that behavioral cues like the ones described in metrics $M_{1-4}$ have a higher chance of transferring between simulation and hardware  than costs. On hardware, once we have evaluated a controller with a particular value of $\phi$, we expect controllers with similar values of $\phi$ to have a similar cost. This roughly splits the cost function landscape, separating points that can potentially walk, and those that cannot, as shown in Figure \ref{fig:dogvsfit}. Suppose we sample an unstable point with a low $\phi$ score and obtain a high cost on hardware. We can then be fairly certain of other unstable points doing poorly as well.
As a result, we can focus on potentially promising points to sample -- making the search more sample efficient and biased towards sampling safe points.

Note that in Equation \ref{eq:dog}, metrics $M_{1-4}$ are weighed equally when summing up. A small but useful addition could be to learn the weights for a 4-dimensional feature transform $[M_1, M_2, M_3, M_4]$ using Automatic Relevance Determination (ARD) \cite{GPsMLBook}. This would enable us to weigh the 4 metrics depending on their importance for a particular task, controller or robot. We leave experimenting with this for future work.

\subsection{Bayesian Optimization with the DoG Transform}

$\phi$ defines a reparameterization of a point from the original space of controller parameters into a 1-dimensional space. We use $\phi$ to define a 1-dimensional kernel that utilizes the Determinants of Gait scores. The functional form of this kernel is the same as Squared Exponential kernel. However, instead of Euclidean distances of points in the original space, we use distances between the DoG scores of the points:
\begin{align}
    k(\pmb{x}_i, \pmb{x}_j) \rightarrow k(\phi(\pmb{x}_i), \phi(\pmb{x}_j)) \\
    k_{DoG}(\pmb{x}_i, \pmb{x}_j) = \sigma_k^2 \exp\Big(- \frac{1}{2 \pmb{l}^2} \|\phi(\pmb{x}_i) - \phi(\pmb{x}_j)\|^2 \Big),
\end{align}
where hyperparameters $\sigma_k^2, \ \pmb{l}^2$ are signal variance and length scale respectively. We refer to $k_{DoG}$ as `\dogkernel' in the following sections.
To speed up calculation of kernel distances during optimization, we pre-calculate $\phi$ for a large grid of points in simulation. We run short simulations of each point/controller, evaluate $\phi$ and store it in a large look-up table.

While the proposed generalized transform $\phi$ can successfully characterize the quality of a gait, large mismatch between a simulator and real-world hardware could still present a challenge. Some controller parameters could yield good gait characteristics in a short simulation, but perform poorly during a longer trial on hardware or simulation. While this issue did not arise during our hardware experiments with the controller described in Section~\ref{sec:raibert_cont}, we anticipate that with a different and higher-dimensional controller such mismatch could become an issue. Hence for our experiments with 50 dimensional virtual neuromuscular controller we explore learning the mismatch and adjusting the kernel accordingly.

We expand the \dogkernel to have one more dimension. This dimension is used to model the anticipated simulation-hardware mismatch with a (separate) Gaussian Process: $g(\pmb{x})\!\!\sim\!\!\mathcal{GP}(0, k_{SE}(\pmb{x}_i,\pmb{x}_j))$. We start with a prior mismatch of zero. For each controller $\pmb{x}_i$ explored during BO, we observe the difference between its DoG score in simulation and on hardware: $d_{\pmb{x_i}}\!=\!\phi_{sim}(\pmb{x_i}) \!-\! \phi_{hw}(\pmb{x_i})$. This difference $d_{\pmb{x_i}}$ becomes a ``training point'' for the GP that is used to model the mismatch. The posterior mean $g_*$ is then computed using standard Gaussian Process regression. This allows us to predict simulation-hardware mismatch for the whole space of controller parameters. The re-parameterization becomes: $\pmb{\phi}^{adj}_{\pmb{x_i}} = \big[ \phi(\pmb{x_i}), g_*(\pmb{x_i}) \big]$,
\begin{align}
    k_{DoG_{adj}}(\pmb{x_i}, \pmb{x_j}) = \sigma^2_k exp\Big( -\frac{1}{2\pmb{l}^2} || \pmb{\phi}^{adj}_{\pmb{x_i}} - \pmb{\phi}^{adj}_{\pmb{x_j}} ||^2 \Big)
\end{align}
Suppose we evaluate controller parameters $\pmb{x_i}$, and in simulation we get walking, but on hardware we get falling. For the next few evaluations, BO with \dogkernel would associate high simulation-based DoG scores with bad performance. In contrast, $k_{DoG_{adj}}$ takes into account the high mismatch (high DoG score in simulation, low on hardware). Consequently, $k_{DoG_{adj}}(\pmb{x_i}, \pmb{x_j})$ would be highest only for $\pmb{x_j}$s that have both similar simulation-based DoG scores and similar estimated mismatch. Hence, points with high simulation-based DoG scores and low predicted mismatch would still be `far away' from the failed $\pmb{x_i}$. They could be sampled if uncertainty about their costs is high.

We present preliminary results of this adjusted \dogkernel in Section \ref{sec:vnmc_expt}.
This formulation is similar in spirit to other approaches, such as \cite{marco2017virtual}. However, while previous work ``mistrusts" all simulation data, our formulation lets us fit a dynamic mismatch function from data. This lets us trust the simulation in some regions, while mistrust it in others.

\section{Atrias robot and controllers}
\label{sec:atrias_cont}
    In this section, we describe our test platform, the ATRIAS robot and the controllers tested in this paper. 

\subsection{ATRIAS robot}
\label{sec:atrias}
Our test platform is CMU's ATRIAS robot (Figure \ref{fig:atrias}), a human sized bipedal robot. The ATRIAS robot was designed so that the inertial properties of the Center of Mass of ATRIAS matched that of humans. The robot weights about $64kg$, with most of its mass concentrated around the trunk. The torso is located about $0.19m$ above the pelvis, and its rotational inertia is about $2.2 kgm^2$. The legs are 4-segment carbon-fiber linkages driven with a point foot, making the legs very light and enabling fast swing movements. The legs are actuated by 2 Series Elastic Actuators (SEAs) in the sagittal plane and a DC motor in the lateral plane. 
Although ATRIAS is capable of 3D walking, in this work, we focus on planar movements around a boom.

\subsection{Feedback based reactive stepping policy}
\label{sec:raibert_cont}
We design a parametrized controller for controlling the CoM height, torso angle and the swing leg as follows:
\begin{align}
    \label{eq:raibert}
    F_x = K_{pt}(\theta_{des} - \theta) + K_{dt}(\dot{\theta}_{des} - \dot{\theta}) \\
    F_z = K_{pz}(z_{des} - z) + K_{dz}(\dot{z}_{des} - \dot{z})\\
    x_p = k(v-v_{tgt}) + C \cdot d + 0.5 \cdot v \cdot T
\end{align}
Here, $F_x$ is the desired horizontal ground reaction force (GRF), $K_{pt}$ is the proportional gain on the torso angle $\theta$ and $K_{dt}$ is the derivative gain on the torso angular velocity $\dot{\theta}$. $\theta_{des}$ and $\dot{\theta}_{des}$ are the desired torso lean and desired torso angular velocity. $F_z$ is the desired vertical GRF, $K_{pz}$ is the proportional gain on the CoM height $z$ and $K_{dz}$ is the derivative gain on the CoM vertical velocity $\dot{z}$. $z_{des}$ and $\dot{z}_{des}$ are the desired CoM height and desired CoM vertical velocity. Both $\dot{\theta}_{des}$ and $\dot{z}_{des}$ are always set to $0$. $x_p$ is the desired foot landing location for the end of swing; $v$ is the horizontal CoM velocity, $k$ is the feedback gain that regulates $v$ towards the target velocity $v_{tgt}$. $C$ is a constant and $d$ is the distance between the stance leg and the CoM; $T$ is the swing time and the term $0.5 \cdot v \cdot T$ is a feedforward term similar to a Raibert hopping policy \cite{raibert1986legged}.

\begin{figure}[t]
\centering
\includegraphics[width=0.35\textwidth]{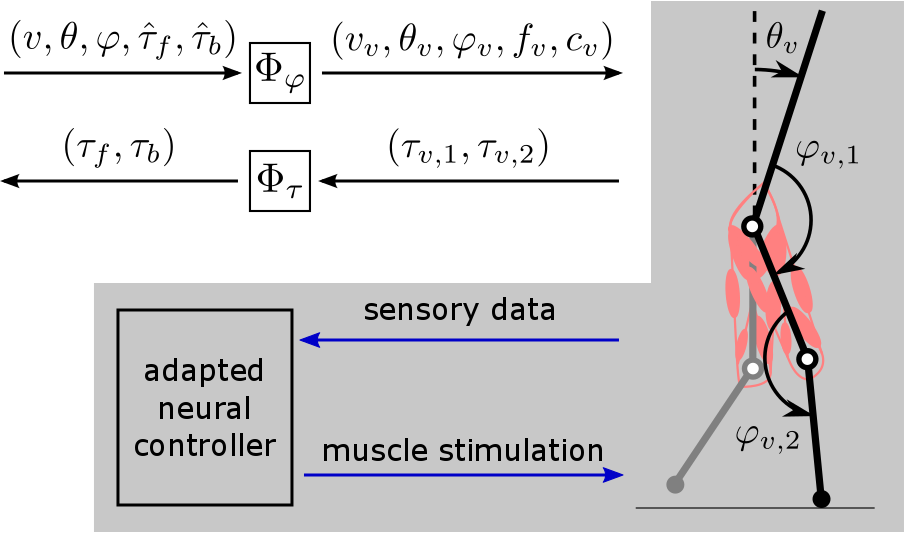}
\caption{\small{Virtual neuromuscular control.
VNMC maps the robot's state, $(v, \theta, \varphi, \hat{\tau}_f, \hat{\tau}_b)$, to virtual measurements required to emulate a neuromuscular model, $(v_v, \theta_v,  \varphi_v, f_v, c_v)$, where $\varphi$ are joint angles, and $f_v$ and $c_v$ are force and contact data of the virtual leg.
The virtual neuromuscular model (in the gray box) outputs virtual joint torques, $(\tau_{v,1}, \tau_{v,2})$, that are mapped to desired robot joint torques, $(\tau_f, \tau_b)$, which are tracked by the SEA controller.}}
        \vspace{-5mm}
\label{fig:VNMC}
\end{figure}

This parametrization results in desired ground reaction forces (GRFs) in stance and a desired foot landing position in swing. In stance, the desired GRFs are then sent to the ATRIAS inverse dynamics model that generates desired motor torques $(\tau_f, \tau_b)$ that realize the GRFs. Details can be found in \cite{wu2014highly}. These desired motor torques are then sent to a low level motor velocity-based feedback loop that generates the desired torques in the robot SEAs.

In swing, we generate a 5th order spline that starts from the current position and velocity of the swing leg, $x_{sw}$ and $\dot{x}_{sw}$ and terminates at the desired foot position $x_{fp}$, with ground speed matching (swing leg is at rest with respect to the ground). This trajectory gives the desired position and velocity of the swing leg, $x_{sw}^*$, $\dot{x}_{sw}^*$, which is translated to desired joint positions and velocities using the robot kinematics. These are then position-controlled by sending a velocity command to the robot SEAs.

This controller assumes no double-stance, swing leg takes off as soon as stance is detected. This leads to a highly dynamic gait, as the contact polygon for ATRIAS in single stance is a point. 
The controller also depends on the desired speed of walking (as this determines the next stepping location). This means that the ``stability" of the controller depends not only on the parameters chosen, but also the desired target speed. We assume that the target speed is provided by the user and is constant in our experiments. 

\begin{enumerate}
    \item 5 dimensional walking controller : In our first set of experiments, we optimized 5 parameters from the above described controller. These were $[K_{pt}, K_{dt}, k, C, T]$. The desired positions and velocities were hand tuned, and so was the feedback on $z$.
    \item 9 dimensional walking controller : In our second set of experiments, we optimized 9 parameters of the above described controller. They were $[K_{pt}, K_{dt}, \theta_{des}, K_{pz}, K_{dz}, z_{des}, k, C, T]$
\end{enumerate}

\subsection{Virtual Neuromuscular Controller for ATRIAS}
\label{sec:VNMC_cont}

We adapt a previously proposed virtual neuromuscular controller (VNMC) \cite{batts2015toward}.
VNMC maps a neuromuscular model to the robot's topology and emulates it to generate desired motor torques, which is sent to the SEA controller (Figure \ref{fig:VNMC}).
The emulated neuromuscular model, which is originally developed to study human locomotion, consists of primarily spinal reflexes, and with appropriate sets of control parameters, it generates diverse human locomotion behaviors \cite{song2015neural} and reacts to a range of external perturbations as observed in humans \cite{song2017evaluation}.

For this study, we adapt the previous VNMC \cite{batts2015toward} by removing some unnecessary biological components while preserving its basic functionalities.
First, the new VNMC directly uses joint angular and angular velocity data instead of estimating it from physiologically plausible sensory data, such as muscle fiber states, when applicable.
Second, most of the neural transmission delays are removed, except the ones utilized by the controller.

The adapted VNMC consists of 50 control parameters, and when optimized using covariance matrix adaptation evolution strategy \cite{hansen2006cma}, it can control ATRIAS to walk on rough terrains with height changes of $\pm$20 cm in planar simulation.

\section{Experiments}
\label{sec:experiments}

In this section we describe our experiments on the ATRIAS hardware and simulation. We test the 5 dimensional controller and the 9 dimensional controller on hardware, both described in Section \ref{sec:raibert_cont}. In addition, we test the 50 dimensional VNMC controller on ATRIAS simulation, described in \ref{sec:VNMC_cont}. We compare BO with the \dogkernel (our method) with BO with Squared Exponential (SE) kernel, which is a commonly used kernel in BO. 

\subsection{Experiments with a 5 dimensional controller}

\label{sec:hdw_5d}
The first set of hardware experiments were conducted on a 5-dimensional controller, described in Section~\ref{sec:raibert_cont}. The target speed profile for these experiments was $0.4 m/s \text{ (15 steps)} - 1.0 m/s \text{ (15 steps)} - 0.2 m/s \text{ (15 steps)} - 0 m/s \text{ (5 steps)}$. The total number of steps before the controller shut off were $50$. The cost function that was optimized was:
\begin{equation}
    \label{eq:hdw_cost}
    cost = 
    \begin{cases}
		100 - x_{fall} , \text{\small{if fall}} \\
		||v_{avg} - v_{tgt}||, \text{\small{if walk}}\\
	\end{cases}
\end{equation}

where $x_{fall}$ is the distance covered before falling, $v_{avg}$ is the average speed per step and $v_{tgt}$ is the target velocity profile. 

We sampled 100 random points on hardware and 10 of them walked for this profile. This means that random sampling has a 1/10 chance of sampling a good point. In simulation, 276 points out of 1000 randomly sampled points walked, implying a 1/4 success rate. This highlights the difference between hardware and simulation, making this a much tougher problem on hardware.

\begin{figure}[t]
\centering
\includegraphics[width=0.375\textwidth]{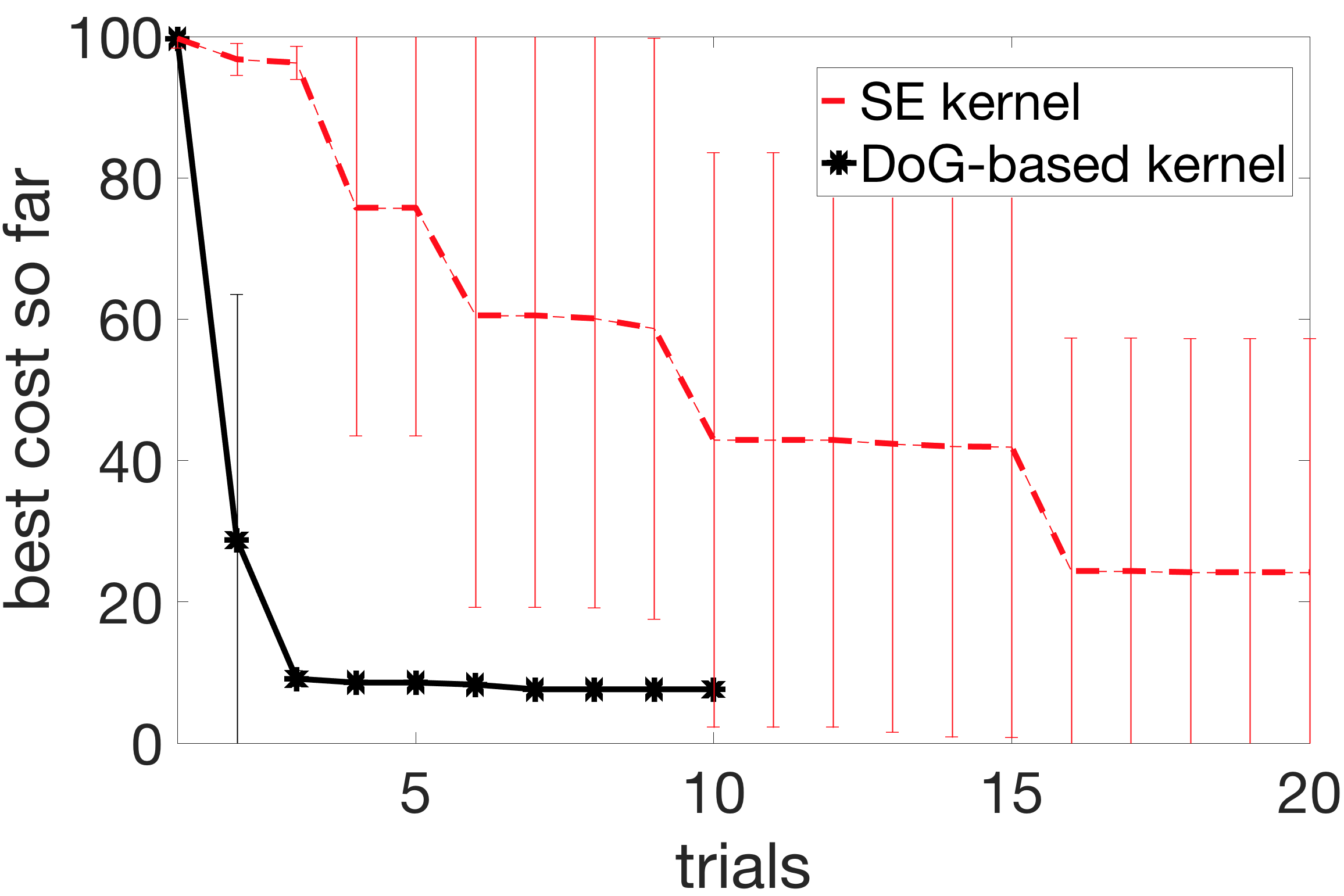}
\caption{\small{BO for 5 dimensional controller on ATRIAS robot hardware. BO with SE finds walking points in 4/5 runs in 20 trials. BO with \dogkernel finds walking points in 5/5 runs in 3 trials.}}
        \vspace{-5mm}
\label{fig:hw_raibert_5d}
\end{figure}

On hardware, we conducted 5 runs of each -- BO with \dogkernel and BO with SE, 10 trials for \dogkernel per run, and 20 for SE kernel. In total, this led to 150 experiments on the robot (excluding the 100 random samples). We also experimented with using fixed vs automatically learned hyperparameters for both kernels. A simple choice of fixed hyperparameters worked well for \dogkernel, while for SE kernel it was better to learn these automatically. DoG was calculated on 20,000 points in simulation, by running $3.5s$ long simulations with target speed of $0.5m/s$.

BO with \dogkernel found walking points in 3 trials in 5/5 runs. BO with SE found walking points in 10 trials in 3/5 runs, and in 4/5 runs in 20 trials. These results can be seen in Figure \ref{fig:hw_raibert_5d}.

\subsection{Hardware experiments with a 9 dimensional controller}
The second set of hardware experiments were conducted on a 9-dimensional controller, described in Section \ref{sec:raibert_cont}. The target speed profile for these experiments was $0.4 m/s \text{ (30 steps)}$. The total number of steps before the controller shut off was $30$. The cost optimized was the same cost as in Equation \ref{eq:hdw_cost}. 

We sampled 100 random points on hardware and 3 of them walked for this speed profile. This means that random sampling has a 1/33 chance of sampling a good point. On the original profile from Section \ref{sec:hdw_5d}, the number of successful points out of 100 were 0, implying a less than 1\% success rate. To keep the problem at hand reasonable, we used simpler target speed profile. In comparison, the success rate in simulation is 8\% for the tougher profile, implying mismatch between hardware and simulation again. 

For this setting, we conducted 3 runs of each BO with \dogkernel and BO with SE, 10 trials for \dogkernel per run, and 10 for SE. In total, this led to 60 experiments on the hardware (excluding the random sampling). 

\begin{figure}[t]
\centering
\includegraphics[width=0.375\textwidth]{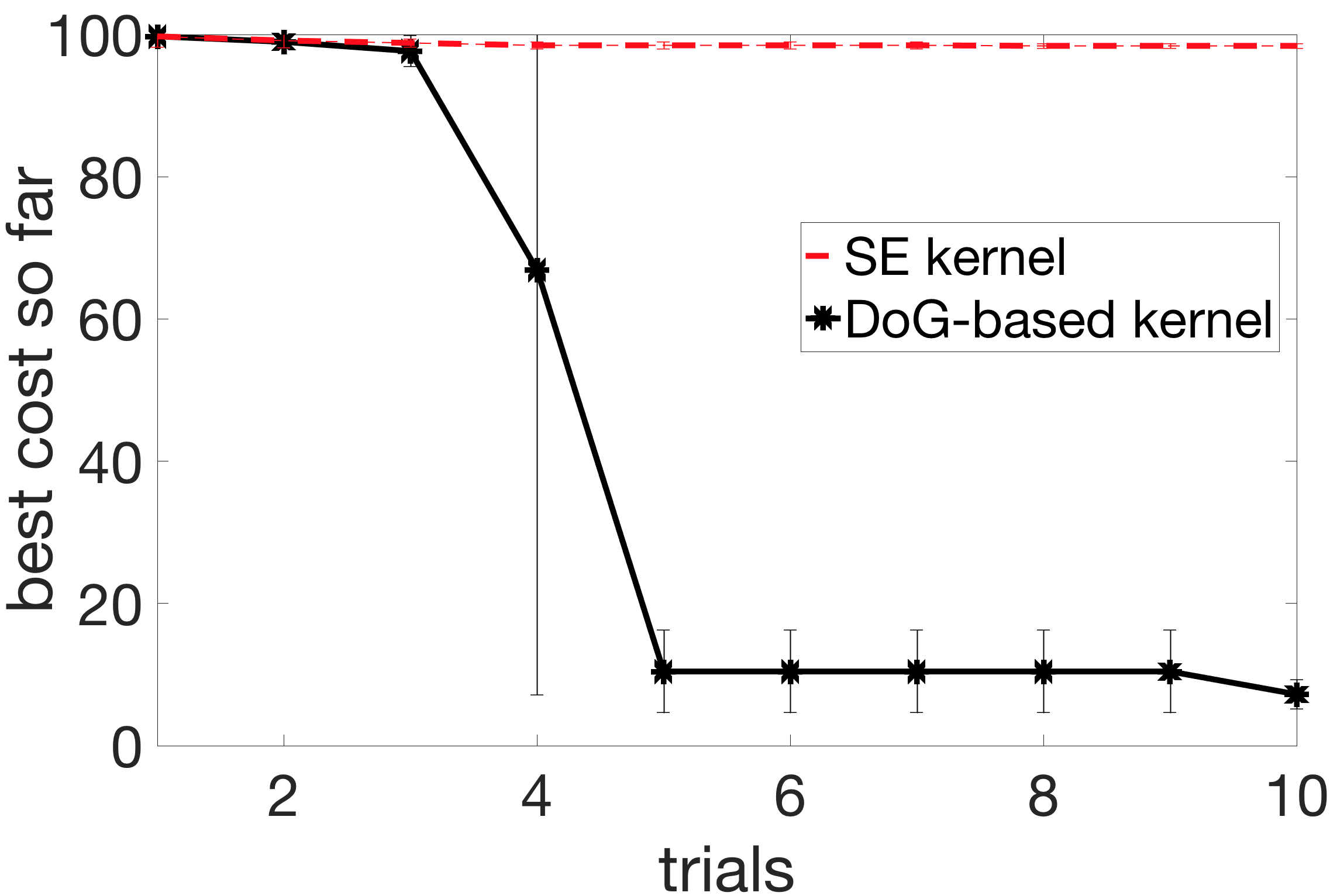}
\caption{\small{BO for 9 dimensional controller on ATRIAS robot hardware. BO with SE doesn't sample any walking points in 3 runs. BO with \dogkernel finds walking points in 5 trials in 3/3 runs.}}
        \vspace{-5mm}
\label{fig:hw_raibert_9d}
\end{figure}

BO with \dogkernel found walking points in 5 trials in 3/3 runs. BO with SE did not find any walking points in 10 trials in all 3 runs. These results can be seen in Figure \ref{fig:hw_raibert_9d}.

Based on these results, we concluded that BO with \dogkernel was indeed able to extract useful information from simulation and speed up learning on hardware.

\subsection{Simulation experiments with a 9-dimensional controller}
\label{sec:exps_sim}
To facilitate further experiments we used ATRIAS simulator~\cite{martin2015robust} with modeling disturbances and different target speed profiles. For these experiments, we use the 9-dimensional controller described in \ref{sec:raibert_cont} as this proved to be a more challenging setting for the ATRIAS hardware. Masses of the robot torso, legs, the boom, as well as inertia of the torso were perturbed randomly by up to 15\% of their original values. This ensured a mismatch between the setting used to generate the kernel and the experimental setting for evaluating its performance, aimed at capturing the discrepancy between hardware and simulation. Note that the kernel was generated on the unperturbed setting, with parameters as described in Section \ref{sec:atrias} for a target speed of $0.5m/s$. The grid size was 100,000 points, and simulations were run for $5s$.

The cost used for these experiments was
\begin{equation}
\label{eq:cost_sim}
cost = 		
    \begin{cases}
		100 - x_{fall} , \text{\small{if fall}} \\
		||v_{avg} - v_{tgt}|| + c_{tr}, \text{\small{if walk}}\\
	\end{cases}
\end{equation}
where $x_{fall}$ is the distance covered before falling, $v_{avg}$ is the average speed per step, $v_{tgt}$ is the target velocity for that step, and $c_{tr}$ captures the cost of transport - calculated by taking a sum of motor torques and normalizing them by a constant. Simulations for evaluating the cost were run for $30s$. Note the addition of the cost of transport in this cost, as compared to \ref{eq:hdw_cost}. $c_{tr}$ needs more than 10 trials to be optimized significantly, and the current low-level motor controllers in Section \ref{sec:raibert_cont} are not designed to reduce $c_{tr}$. Hence, its not considered in the hardware experiments.
\begin{figure}[t]
\centering
\includegraphics[width=0.37\textwidth]{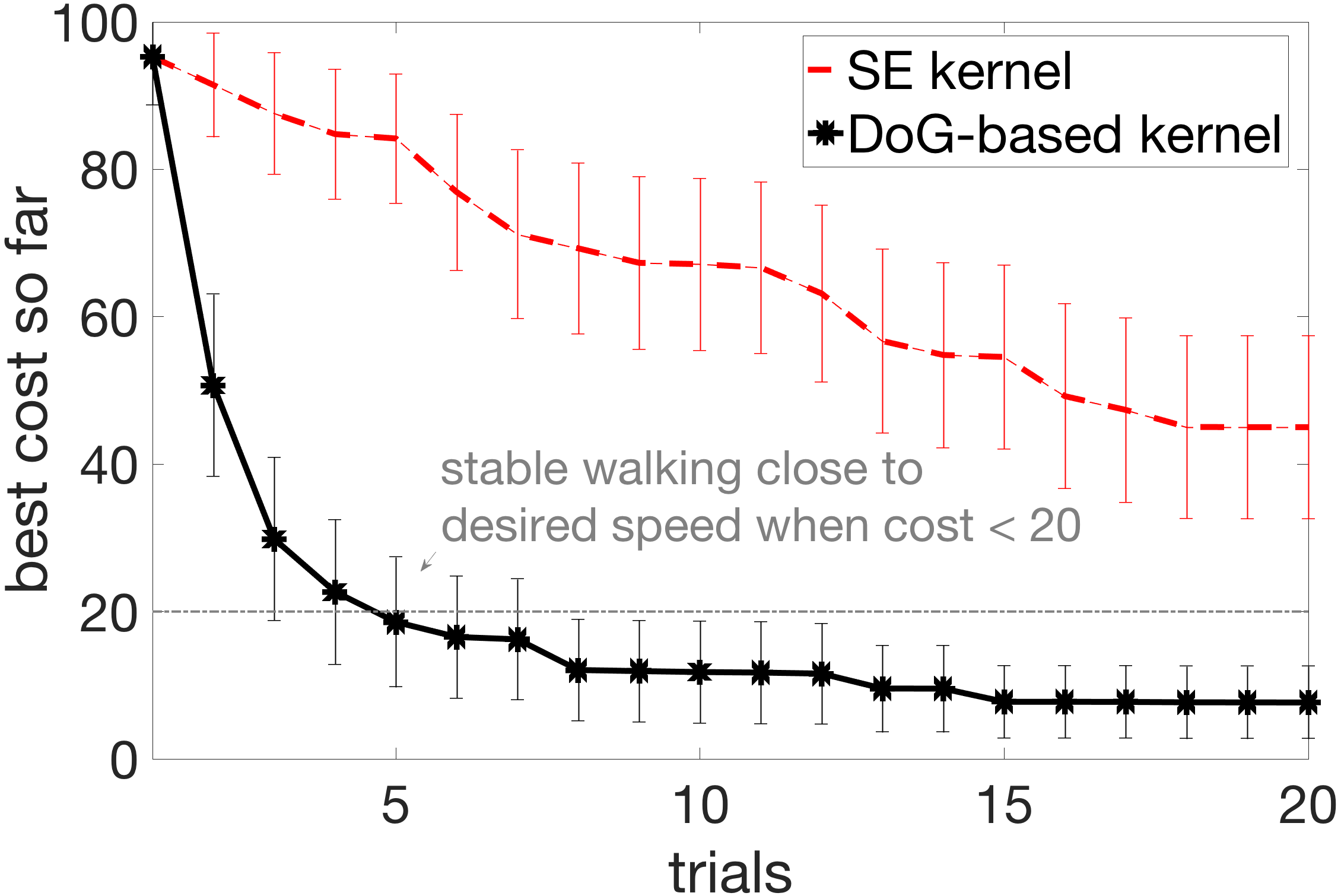}
\caption{\small{Further experiments using ATRIAS simulator: BO for ``speed-up-down'' target speed profile on robot model with mass and inertia differences
(mean over 50 runs; 95\% confidence intervals).}}
\label{fig:sim_9d_disturb}
\vspace{-7mm}
\end{figure}
Figure~\ref{fig:sim_9d_disturb} illustrates BO on a simulated model with mass and inertia differences. The target was to start walking at 0.4m/s, then speed up to 0.6m/s, then 1.0m/s, slow down to 0.6m/s, then walk at 0.2m/s. \dogkernel was collected using an unperturbed model with a target speed of 0.5m/s, and yet it performed very well on this more challenging setting. After 20 trials, 96\% of BO runs using the \dogkernel found a stable walking solution, compared to 56\% of the runs using an SE kernel. The average cost of the walking solutions was also improved: lower by $\approx$30\% when using DoG vs SE kernel.

These experiments suggest that \dogkernel is able to offer improvement for the settings different from the one used to generate it. This improvement is robust to both the deviations of the robot model/hardware parameters as well as desired walking speed profiles.

\subsection{Experiments on a 50 dimensional Virtual Neuromuscular Controller}
\label{sec:vnmc_expt}
To explore the possibility of using \dogkernel with high-dimensional controllers we experimented with Virtual Neuromuscular Controller (VNMC) described in \mbox{Section~\ref{sec:VNMC_cont}}.

VNMC does not start from rest, and needs an initial velocity. In previous work this has been emulated by either giving simulations initial speeds, or by giving a push. These are either un-realizable or unreliable on hardware. To overcome this problem, we start the VNMC with a \mbox{5-dimensional} walking controller (described in Section \ref{sec:atrias_cont}, parameters hand-tuned and fixed). Once the robot has taken 10 steps with this controller, the control is switched to VNMC. 

To construct \dogkernel for this controller we collected 250,000 points from 7-second simulations. The DoG scores were computed after switching to the VNMC (so after first 10 steps). Searching in 50 dimensional space could be completely intractable if the search region is too wide. Usually enough domain knowledge is available to confine the search to a reasonably manageable region. We tried to find an initial point that walked 3-4 steps before falling in simulation (this point still had a high cost of $\approx\!\!93$). This point became the ``center" of our search space, and we searched in a hyper-cube of size $[0.75, 1.25]$ in each dimension around this point. So, with initial point $x_0$, the search space was $[0.75 \cdot x_0 , 1.25 \cdot x_0]$. With these boundaries, 4\% of points sampled randomly were walking.

\begin{figure}[t]
\centering
\includegraphics[width=0.37\textwidth]{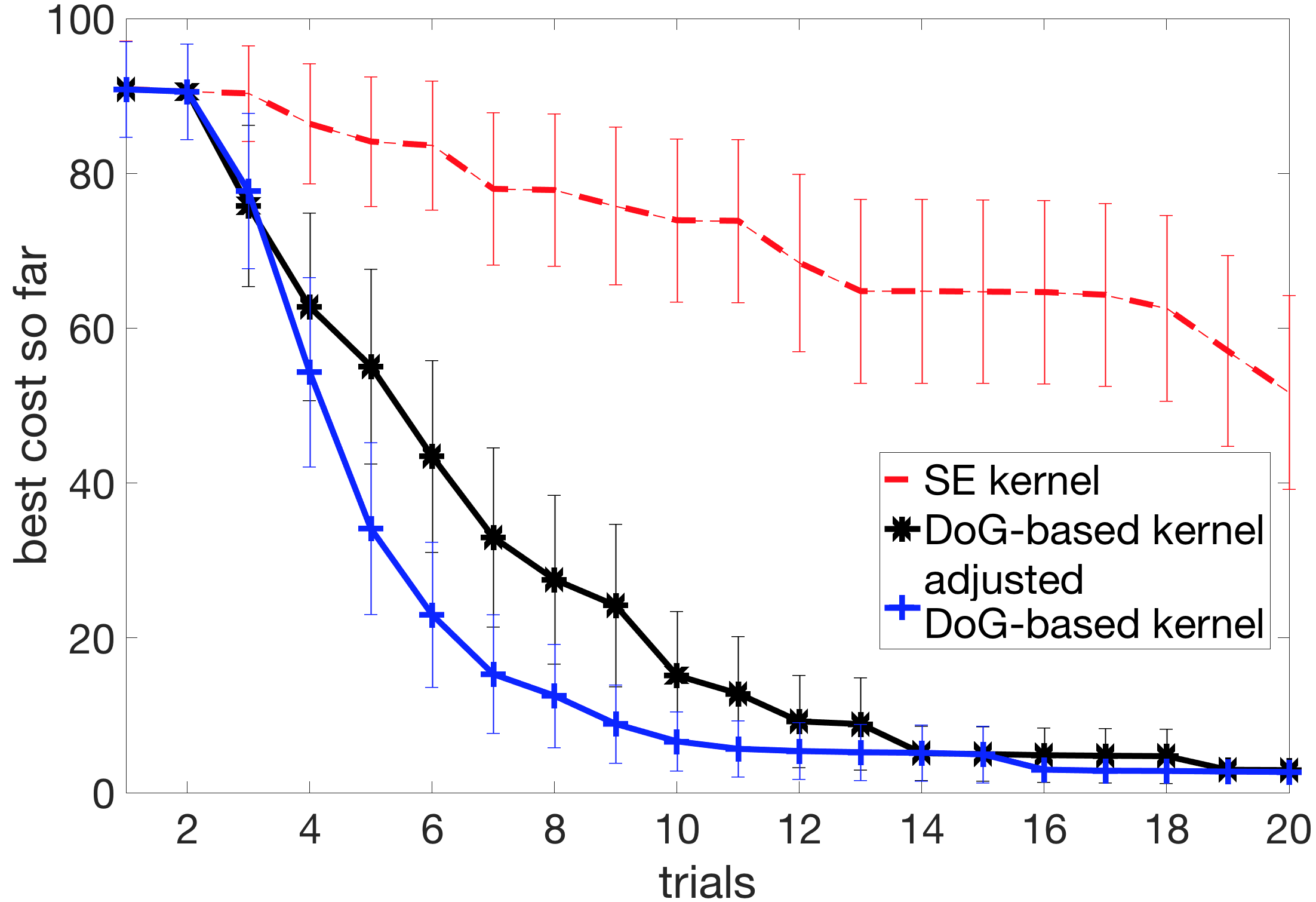}
\caption{\small{BO with Virtual Neuromuscular Controller.}}
\label{fig:sim_NM_Atrias_50d}
\vspace{-7mm}
\end{figure}

Figure \ref{fig:sim_NM_Atrias_50d} shows results of Bayesian Optimization with SE, \dogkernel and adjusted \dogkernel. During optimization simulations were run for $30$ seconds. We used the same cost function as described in the previous section (equation~\ref{eq:cost_sim}).

In 50-dimensional control, the mismatch between long and short simulations becomes apparent. 
For the 5-dimensional and 9-dimensional controllers, the performance during short simulations usually predicted whether $30s$ simulations would be successful. That is, points that walk for 5s would walk for 30s. However, this is not true for the 50-dimensional controller. Since this controller is capable of much richer behaviors, if a point is not in a limit cycle before the end of a short simulation, it can lead to a range of behaviours later. As a result, we noticed an improvement when using adjusted \dogkernel described in Section~\ref{sec:dog}. While DoG is still very competitive and finds walking points in 100\% of the runs by 20 trials, the adjusted DoG with mismatch has an advantage. It reaches the same optimum found by DoG faster. 

\begin{figure*}
\centering
\includegraphics[width=0.9\textwidth]{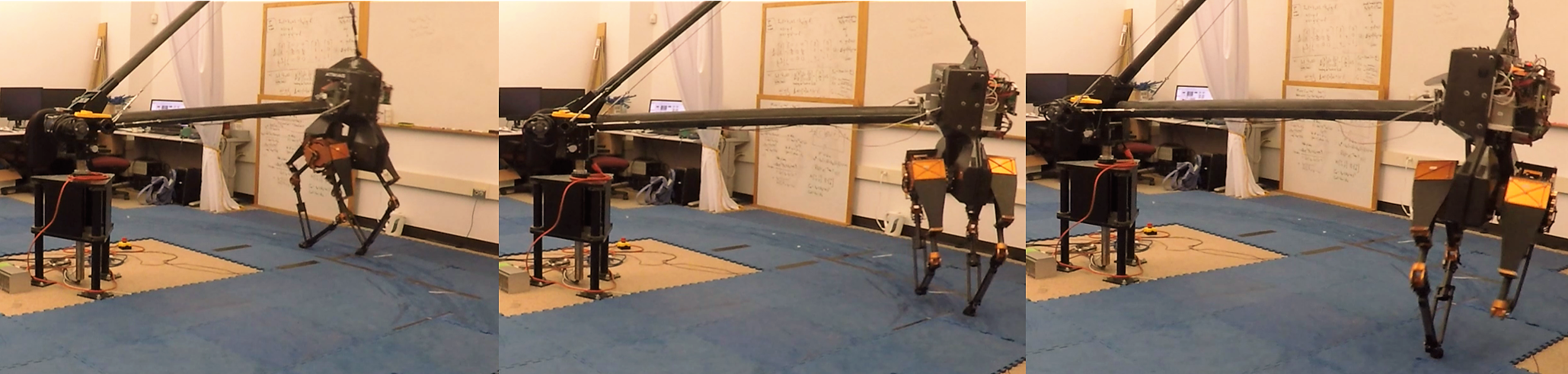}
\caption{\small{A time lapse of ATRIAS walking around the boom during a run of \dogkernel.}}
\label{fig:bo_runs_atrias_hw_slides}
\end{figure*}

The 50-dimensional controller has not been fully implemented to work on hardware yet due to lack of time. However, our experiments so far seem promising and we are working towards a hardware implementation. 
To anticipate potential mismatch between simulation and hardware, we tested it on slightly perturbed initial conditions for the VNMC. The different conditions were aimed to replicate issues likely to be seen on hardware. The starting states for VNMC would differ slightly each time, since they would depend on the state of the robot after the 5-dimensional initiating controller has finished. Both DoG and adjusted DoG were robust to slight changes in initial condition. 
We hope to test adjusted \dogkernel more in the future in this challenging setting that could be sensitive to simulation-hardware mismatch.

\section{Conclusion}
\label{sec:conclusions}
In this work, we proposed a general locomotion kernel that can be used to learn controllers on hardware efficiently. Our hardware experiments on a 5-dimensional and 9-dimensional controller show that indeed \dogkernel transfers important features of walking controllers between simulation and hardware. We also do simulation experiments on a 50-dimensional controller and the results seem promising. We are working on testing this on hardware. 

Using simulation to guide hardware experiments raises some important questions. For example, how can we determine the important features for this transfer? How can we determine their relative importance? How can we propagate the encountered differences between simulation and hardware? We use domain knowledge to extract important features, but one could also use learning to do so \cite{antonova2017deep}. We suggest using Automatic Relevance Determination to adjust relative importance and a mismatch map to propagate differences between simulation and hardware. Both of these methods remain to be tested extensively in simulation and on hardware, but seem promising from initial results. We plan to continue experimenting with these to come up with a robust kernel that can be used for complex walking robots and controllers.




\bibliographystyle{IEEEtran}
\bibliography{references}

\end{document}